%% file: neurips_workshop.tex
\documentclass{article}

\PassOptionsToPackage{square, numbers, sort&compress}{natbib}

\usepackage[final]{neurips_2023}

\usepackage[utf8]{inputenc} 
\usepackage[T1]{fontenc}    
\usepackage{hyperref}       
\usepackage{url}            
\usepackage{booktabs}       
\usepackage{amsmath}   
\usepackage{amssymb}   
\usepackage{amsfonts}       
\usepackage{nicefrac}       
\usepackage{microtype}      
\usepackage{xcolor}         
\usepackage{graphicx}
\usepackage{algorithm}
\usepackage{algpseudocode}
\usepackage{multirow}
\usepackage{caption}
\usepackage{subcaption}
\usepackage[capitalise]{cleveref}
\usepackage{array}
\newcolumntype{x}[1]{>{\centering\arraybackslash\hspace{0pt}}p{#1}}

\title{Decoding Data Quality via Synthetic Corruptions: Embedding-guided Pruning of Code Data}

\author{%
Yu Yang$^{1,2}$\thanks{Work done during internship at Meta.} \\\texttt{yuyang@cs.ucla.edu} \And Aaditya K. Singh$^{2}$ \\ \texttt{aaditya.singh.21@ucl.ac.uk} \And Mostafa Elhoushi$^{2}$ \\ \texttt{melhoushi@meta.com} \And Anas Mahmoud$^{2}$ \\ \texttt{nas.mahmoud@mail.utoronto.ca} \And \textbf{Kushal Tirumala}$^{2}$ \\ \texttt{ktirumala@meta.com}  \And \textbf{Fabian Gloeckle}$^2$ \\ \texttt{fgloeckle@meta.com} \And \textbf{Baptiste Rozière}$^2$ \\ \texttt{broz@meta.com}
 \And \textbf{Carole-Jean Wu}$^2$ \\ \texttt{carolejeanwu@meta.com}  \And \textbf{Ari S. Morcos}$^{3}$\thanks{Work done at Meta.}  \And \textbf{Newsha Ardalani}$^2$ \\ \texttt{new@meta.com}\AND
$^1$UC Los Angeles \quad $^2$FAIR at Meta \quad $^3$DatologyAI\\
}

\begin{document}

\maketitle

\begin{abstract}
Code datasets, often collected from diverse and uncontrolled sources such as GitHub, potentially suffer from quality issues, thereby affecting the performance and training efficiency of Large Language Models (LLMs) optimized for code generation. Previous studies demonstrated the benefit of using embedding spaces for data pruning, but they mainly focused on duplicate removal or increasing variety, and in other modalities, such as images. Our work focuses on using embeddings to identify and remove ``low-quality'' code data. First, we explore features of ``low-quality'' code in embedding space, through the use of synthetic corruptions. Armed with this knowledge, we devise novel pruning metrics that operate in embedding space to identify and remove low-quality entries in the Stack dataset. We demonstrate the benefits of this \textit{synthetic corruption informed pruning (SCIP)} approach on the well-established HumanEval and MBPP benchmarks, outperforming existing embedding-based methods. Importantly, we achieve up to a 3\% performance improvement over no pruning, thereby showing the promise of insights from synthetic corruptions for data pruning.
\end{abstract}

\section{Introduction}
\input{sections/1_intro}

\input{sections/3_method}

\section{Pruning Low-quality Data for More Efficient Training}
\input{sections/4_experiments}

\input{sections/5_conclusion}

\bibliography{iclr2024_quality}
\bibliographystyle{unsrt}

\newpage
\appendix
\input{sections/6_appendix}

\end{document}

%% file: sections/1_intro.tex
Machine learning, and in particular Large Language Models (LLMs), are transforming a wide range of industries. Their capabilities extend even to specialized tasks like code generation and medical diagnostics, thus amplifying their societal and economic impact \citep{eloundou2023gpts}. In this race for higher performance, some training datasets have swelled to petabyte size, sourced from extensive repositories like the Common Crawl. While significant effort has gone into optimizing the computational aspects of training LLMs, such as hardware acceleration and algorithmic improvements \citep{Dao2022FlashAttentionFA}, the question of data efficiency is still relatively under-explored. Data efficiency is not merely a computational concern but is intrinsically tied to the quality of the training data. The use of large, but ineffective, datasets can result in protracted training times, higher energy consumption, and ultimately, models that are expensive to deploy and maintain \cite{sorscher2022beyond}.

Code datasets, usually compiled from diverse, open-source platforms like GitHub, are often riddled with inconsistencies, errors, or low-quality code snippets. These issues not only undermine the model's final performance but also affect the efficiency and effectiveness of the training process. The presence of such low-quality data essentially ``pollutes'' the learning environment, leading to suboptimal results. Therefore, improving data quality is not merely an ancillary task but a fundamental requirement for achieving the full potential of code-generating LLMs. A recent study \cite{gunasekar2023textbooks} showcased the benefits of so-called ``textbook-quality'' data in enhancing model efficiency for code-generation tasks. However, their strategy relies heavily on generating closed-source data with GPT-3.5 and then filtering it based on GPT-4 \cite{openai2023gpt4} predictions, both of which are proprietary models, thus making this approach less accessible for many researchers due to high costs and difficulty of reproducibility. Furthermore, another study \cite{shumailov2023curse} highlighted potential issues with training on generated outputs. This emphasizes the need for open-source techniques to identify valuable data in existing, large-scale, natural corpora.

Building upon these identified challenges and gaps in existing research, we focus on easy-to-use, accessible pruning methods for the large open-source Stack dataset \cite{Kocetkov2022TheStack}.
To this end, we take inspiration from recent approaches to data pruning in the domains of image \citep{sorscher2022beyond} and multimodal models \citep{abbas2023semdedup}, which make use of pre-trained embedding spaces to identify useful or duplicate data, to keep or prune, respectively. 
In the hitherto unexplored domain of code, we introduce synthetic corruption informed pruning (SCIP): First, we identify what constitutes ``low-quality'' data in embedding space through controlled corruption of existing data, and find that corrupted code tends to reside in smaller clusters and often be farther from cluster centroids. Then, we introduce a pruning strategy, based on these insights, that ranks data points based on their cluster size and distance to the nearest centroid, aiming to remove a predefined fraction of the data. Using these embedding-based methods for pruning low-quality code, we
demonstrate improvements in performance and training efficiency on widely used benchmarks \citep{chen2021evaluating, austin2021program}. 



%% file: sections/3_method.tex
\section{What Does Low-Quality Mean for Code Data?}

\begin{figure}
    \centering
    \includegraphics[width=\textwidth]{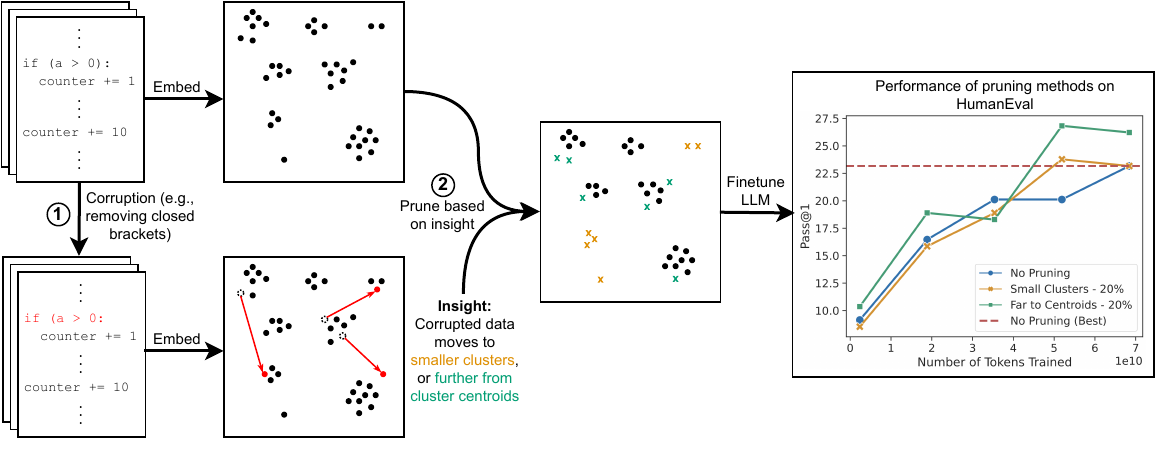}
    \caption{Schematic of SCIP. First, we synthetically corrupt code data, which tends to move code embeddings to smaller clusters or further from cluster centroids. Then, we use this insight to propose a new pruning metric, resulting in improved training efficiency and better end performance.}
    \label{fig:approach}
    \vspace{-1.5em}
\end{figure}

\subsection{Definition of Low-Quality Data}
Let \( \mathcal{D} \) be the original dataset, \( \mathcal{Q} \subseteq \mathcal{D} \) be a subset, and \( \mathcal{D}_{\text{test}} \) be the test set. Let \( x_{\text{test}, i} \) be the \(i\)-th test example in \( \mathcal{D}_{\text{test}} \). First, we define a general metric \( M \), which could potentially be \( \text{pass@k} \) \citep{chen2021evaluating} or any other quality metric. We then define \( M(\theta(\mathcal{D}), \mathcal{D}_{\text{test}}) \) as the expectation of a particular metric (for example, \( \text{pass@k}_i \)) over all \( x_{\text{test}, i} \) in \( \mathcal{D}_{\text{test}} \) when training on dataset \( \mathcal{D} \) with model parameters \( \theta \):

\[
M(\theta(\mathcal{D}), \mathcal{D}_{\text{test}}) = \mathbb{E}_{x_{\text{test}, i} \in \mathcal{D}_{\text{test}}}[\text{pass@k}_i]
\]

The set \( \mathcal{Q} \) is defined as ``low-quality'' if the following inequality holds:

\[
M(\theta(\mathcal{D}), \mathcal{D}_{test}) < M(\theta(\mathcal{D} \setminus \mathcal{Q}), \mathcal{D}_{test})
\]

In simpler terms, \( \mathcal{Q} \) is considered ``low-quality'' data if removing it from \( \mathcal{D} \) improves the score of the general metric \( M \) on \( \mathcal{D}_{\text{test}} \).

\subsection{SCIP: Two-Step Framework for Identifying Low-Quality Data}
To systematically identify low-quality data, we propose a two-step framework, illustrated in Figure \ref{fig:approach}. The first step involves the creation of data with known errors, serving as markers for low-quality data. From this first step, we gather insights on how corruption affects embeddings (obtained with a pretrained model), and use this knowledge to prune data with similar embedding properties. 

\paragraph{Synthetic Corruption Generation}

To identify and prune ``low-quality'' code data, it's important to understand its possible forms. We consider two main domains: syntax errors and content errors. Synthetic corruption has the benefit of creating \textit{matched} pairs of higher and lower quality data, making it more controlled than alternative approaches which could be confounded by style.

\begin{itemize}
    \item \textbf{Data with Syntax Errors:} Syntax errors are clear indicators of bad code, preventing a file from executing successfully. Such issues can be as common as unmatched parentheses or as nuanced as referencing undeclared variables. To intentionally introduce these errors for the sake of our experiments, we employ two main corruptions: removing closed brackets (specifically, \texttt{`)'}, \texttt{`]'}, \texttt{`\}'}) and renaming variables to syntactically invalid names.
    \item \textbf{Data with Content Errors:} Although such code may run without immediate issues, its output might diverge from the intended result due to underlying logical errors. To simulate this, we either alter conditional operators (through negation) or offset array indices (changing \texttt{`i'} to \texttt{`i+1'}) to disrupt data access patterns. 
\end{itemize}

More specifics can be found in \cref{sec:manipulation}. Through these synthetic corruptions, we ensure a systematic introduction of both syntax and content errors, aiding in a more comprehensive identification of ``low-quality'' data. By focusing on a representative sample of errors, we effectively set the stage for the next step: identifying and pruning ``low-quality'' data in large-scale datasets.

\begin{figure}[t]
\begin{center}
\includegraphics[width=0.9\textwidth]{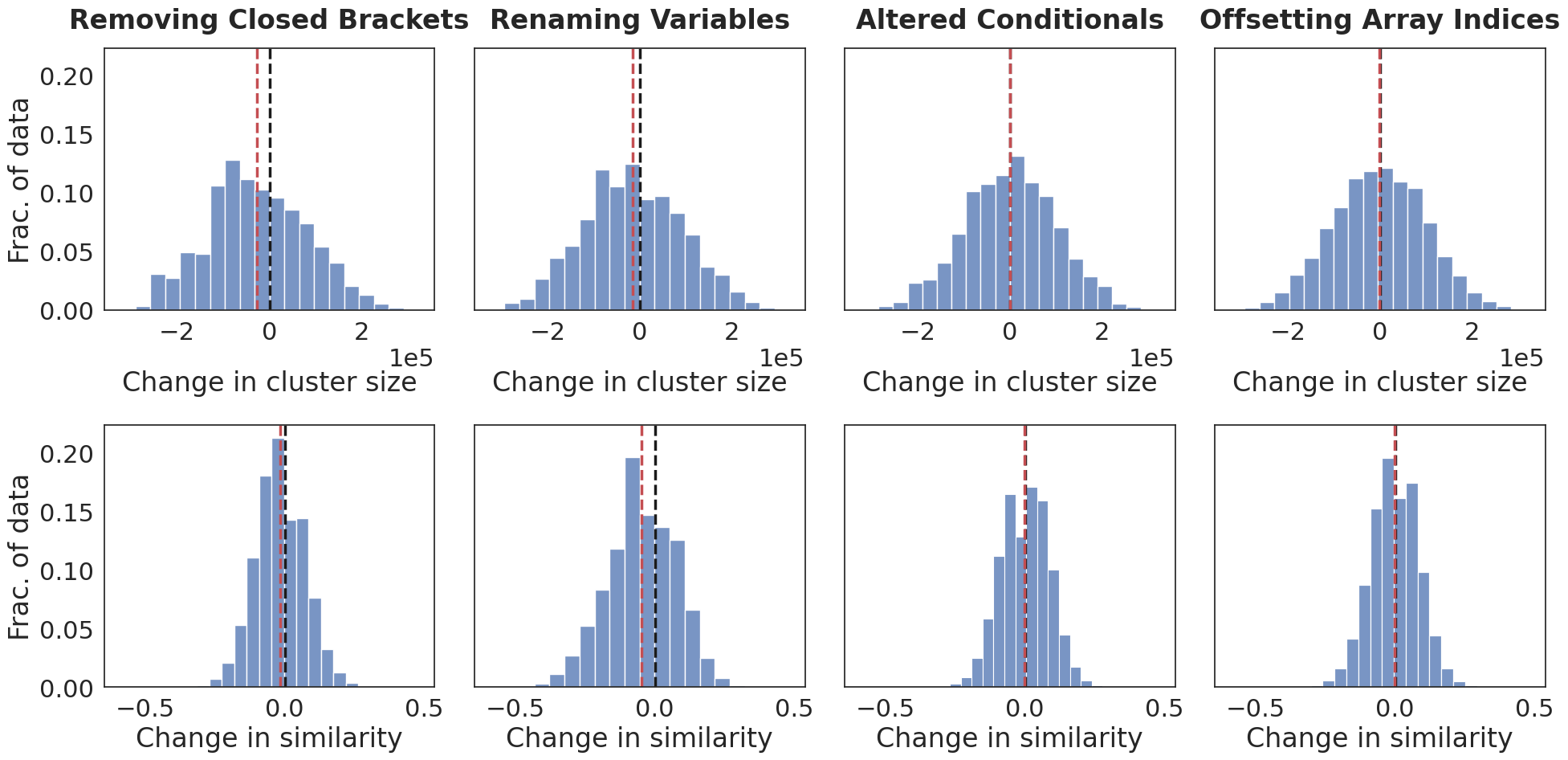}
\end{center}
\caption{Corrupted data tends to reside in smaller clusters (top row) and farther from centroids (bottom row) when compared to the original, uncorrupted data. The effects are more pronounced for syntax errors (left two columns) as compared to content errors (right two columns). Red dotted line indicates mean, black dotted line indicates 0. More details and analysis can be found in \cref{sec:corruption_deepdive}.}
\vspace{-1.5em}
\label{fig:manip_change}
\end{figure}

\paragraph{Data Pruning Informed by Synthetic Corruptions}
In the embedding space of a pre-trained code embedding model, StarEncoder \citep{li2023starcoder}, we see that synthetic corruption exhibits a distinct change: corruption moves points to smaller clusters or further out from centroids, as compared to the original, uncorrupted code (\cref{fig:manip_change}).
These insights shape our pruning strategy. By focusing on data in smaller clusters and distant from centroids, we aim to efficiently identify and remove low-quality data from the original dataset. A formal version of the algorithm, with pseudocode can be found in \cref{sec:alg}.

%% file: sections/4_experiments.tex
\subsection{Experiment Setup}

\textbf{Dataset.} Our experiments utilize the Stack v1.1 dataset \citep{Kocetkov2022TheStack}, which is sourced from GitHub repositories published from 2015 to 2022, and specifically designed for code generation tasks. Although the dataset includes code from 358 different programming languages, we narrow our focus solely to Python to ensure a more controlled study. This results in a dataset of 12.6M files and 20.4B tokens.

\textbf{Model and Training Details.} Following the methodology of the current state-of-the-art open-source model, Code Llama \citep{roziere2023code}, we fine-tune a 1.5B LLaMA \citep{touvron2023llama} model instead of training from scratch. The model has 48 layers, 24 heads per layer, and inner dimension of 1536.
All experiments are run on 32 NVIDIA A100 GPUs with fully-sharded data parallel \citep{FairScale2021}. We use a learning rate of 3e-4, a batch size of 576, a sequence length of 2048, and train for 56,000 steps ($\sim$67B tokens). 

\subsection{Evaluation}
Our evaluation employs two well-established benchmarks in the code generation field: HumanEval \citep{chen2021evaluating} and MBPP \citep{austin2021program}. The primary metric for evaluation across these benchmarks is ``pass@k,'' which measures the percentage of test cases that are correctly solved within the top-k generated code snippets. For baselines, we compare to no pruning, random pruning (averaged over 3 seeds), and three other pruning methods using embeddings, based on prior work in other modalities: SSL-prototypes \citep{sorscher2022beyond}, SemDeDup \citep{abbas2023semdedup}, and D4 \citep{tirumala2023d4}. Additional details can be found in \cref{sec:app-eval}.

\subsection{Results}

In \cref{tab:result}, our proposed methods -- pruning data that are ``Far from Centroid'' and within ``Small Clusters'' -- yield clear performance improvements on HumanEval and MBPP, respectively. However, better performance on one benchmark often comes at the expense of the other, perhaps due to the different natures of these tasks.
Motivated by the strong performance of our two suggested methods, we experimented with a combined method: first pruning files from small clusters, then files far from centroids, with the ratio between these defined by a parameter $\alpha$. We found that $\alpha=0.8$ performed best (see \cref{sec:alg}). Impressively, this combined method achieves the best performance of all methods tried on HumanEval, a full 3\% above no pruning and better than all prior work on embedding-based pruning, while also remaining competitive with no pruning on MBPP. 

We also observe in \cref{fig:approach} that ``Far from Centroid'' and ``Small Clusters'' both achieve an efficiency speedup (both methods achieve the baseline pass@1 rate in fewer training steps). Further insights into the qualitative attributes of pruned data are presented in \cref{fig:examples}."

\begin{table}[t]
    \caption{Pass@1 performance on HumanEval and MBPP for different pruning methods with 20\% files pruned.}
    \centering
    \resizebox{\textwidth}{!}{
    \begin{tabular}{lcccccccc}
    \toprule
      &  No & Random & SSL & \multirow{2}{*}{SemDeDup} & \multirow{2}{*}{D4} & \textbf{Small} & \textbf{Far from} & \textbf{Combined}\\
      &  pruning & Pruning & Prototype &  &  & \textbf{Clusters} & \textbf{Centroids} & \textbf{Small+Far} \\
    \midrule
    HumanEval & {25.0\%} & 24.0\% & 23.8\% & 20.7\% & 23.2\% & 23.2\% & \underline{26.8\%} & \textbf{28.0\%} \\
    MBPP & \underline{33.4\%} & 31.9\% & 32.2\% & 32.4\% & 31.2\% & \textbf{35.0\%} & 30.8\% & 33.0\% \\
    \bottomrule
    \end{tabular}
    }
    \vspace{-0.5em}
    \label{tab:result}
\end{table}

%% file: sections/5_conclusion.tex
\section{Conclusions}

We introduce SCIP, a systematic method to identify and remove ``low-quality'' code data from large datasets. Building on the insights of the value of high-quality data presented in earlier studies \citep{gunasekar2023textbooks}, our work goes further by offering accessible, open-source, and cost-effective pruning techniques through the use of embedding spaces. We go beyond prior work in embedding-based pruning \cite{sorscher2022beyond, abbas2023semdedup,tirumala2023d4} by motivating heuristics through identification of ``low-quality'' data via synthetic corruptions:
we systematically create code discrepancies, both in syntax and content, 
to understand their influence on the embedding space. Our findings reveal that syntax errors lead to significant shifts away from cluster centroids and into smaller clusters. Leveraging these observations, we designed pruning methods that consider both distances to centroids and cluster sizes to effectively identify and remove low-quality data. Applying these pruning methods leads to better performance on code generation benchmarks, showing the promise of insights from synthetic corruptions for improving pruning techniques.

More broadly, our results underscore the significance of rigorous data curation. Beyond just code, more rigorously examining ``low-quality'' data could lead to more informed pruning techniques. Similar to how code can have both syntax and content discrepancies, natural language data too can have structural (e.g., grammatical) and semantic (e.g., factually incorrect) anomalies. In future work, the strategies and methodologies established here of using synthetically corrupted data as a pruning signal could be extended and adapted to general natural language datasets, ensuring models trained on them produce more accurate, reliable, and coherent outputs.

\section*{Acknowledgments}

We would like to sincerely thank Jack Lanchantin for the insightful discussions, and Shubham Toshniwal, Koustuv Sinha, and Alberto Bietti for generously sharing their valuable insights drawn from their previous research.


%% file: sections/6_appendix.tex
\section{Related Work}

\input{sections/2_related}

\section{Synthetically Corrupted Data}
\label{sec:manipulation}

\subsection{Creation}
To effectively identify and prune ``low-quality'' code data, it's important to understand its possible forms. We categorize them into two main domains: syntax errors and content errors.

\paragraph{Data with Syntax Errors}
Syntax errors are clear indicators of problematic code, preventing a code snippet from executing successfully. Such issues can be as common as unmatched parentheses or as nuanced as referencing undeclared variables. To intentionally introduce these errors for the sake of our experiments, we employ two main corruptions: 

\begin{enumerate}
    \item \textbf{Removing Closed Brackets:} By omitting closing brackets, including parentheses \texttt{)}, square brackets \texttt{]}, and curly braces \texttt{\}}, from code, we introduce errors that would halt execution. For example, the code segment \texttt{for i in range(10):} might be changed to \texttt{for i in range(10:}. 
    \item \textbf{Renaming Variables:} Altering variable names at random intervals ensures that they no longer match their original declarations. For instance, a variable declared as \texttt{counter = 0} might be used later in the code as \texttt{counter += 1}, which we would change to \texttt{ctr += 1}, creating a reference to an undeclared variable.
\end{enumerate}

\paragraph{Data with Content Errors}
Although such code may run without immediate issues, its output might diverge from the intended result due to underlying logical errors. To simulate this, we adopt two principal corruptions:

\begin{enumerate}
    \item \textbf{Altered Conditionals:} Switching common relational operators alters the flow of code without introducing blatant syntax errors. For example, conditions like \texttt{if a == b:} can be transformed to \texttt{if a != b:}.
    \item \textbf{Offsetting Array Indices:} Adjusting indices by adding or subtracting a unit disrupts data access patterns. A line such as \texttt{value = array[i]} might become \texttt{value = array[i+1]}, potentially leading to unintended behavior or out-of-bound errors.
\end{enumerate}
Importantly, we note that synthetic corruption yields matched pairs of good (original) and bad (corrupted) code. We corrupt and embed each document in the Stack, with one of the above corruptions at a time. Note that, future work could consider drawing insights from a smaller percentage of corrupted data, but we opted for the full datasets for simplicity. On the whole, this clustering step is relatively inexpensive compared to model training, taking about 2\% of the time. In the next section, we look at the effects of these corruptions in embedding space.

\subsection{Effects of Synthetic Corruptions}
\label{sec:corruption_deepdive}

Building off prior work \cite{sorscher2022beyond, abbas2023semdedup, tirumala2023d4} on embedding-based pruning, we started by clustering the embeddings of our original, unperturbed data. On a single GPU, this clustering step takes on the order of 10 minutes. We used $k=100$ clusters, identified using the K-means algorithm with cosine similarity as the metric. When corrupting a file, we then consider the following two things:
\begin{enumerate}
    \item Distance to its cluster centroid before and after corruption
    \item Size of new cluster if the corrupted file lies in a different cluster. If so, the size difference between the two clusters (in terms of \# of files)
\end{enumerate} 

Our main results are presented in \cref{fig:manip_change}. For visualization pruposes, we excluded all points with negligible change (points that stay in the same cluster or points whose distance to centroid changes by less than 0.01). The fraction of file-pairs remaining after this exclusion is presented in \cref{tab:excluded}. From \cref{tab:excluded}, we can see that content corruptions (right two columns) lead to a way smaller fraction of file pairs that show significant changes in embedding space, which partially explains the weaker signal we see in \cref{fig:manip_change} for these corruptions. On the other hand, syntactical corruptions have effects on most files, especially leading to many cluster changes.

\begin{table}[t]
    \caption{Fraction of files pairs that changed non-negligibly after each corruption.}
    \centering
    \begin{tabular}{p{1.3in}x{1in}x{0.6in}x{0.7in}x{1in}}
    \toprule
      & Removing Closed Brackets & Renaming Variables & Altered Conditionals & Offsetting Array Indices \\ 
    \midrule
    Changed cluster & 0.77 & 0.89 & 0.02 & 0.23 \\
    \hline
    Changed distance from cluster centroid & 0.49 & 0.57 & 0.02 & 0.20 \\
    \bottomrule
    \end{tabular}
    \label{tab:excluded}
\end{table}

\section{Algorithm}
\label{sec:alg}
Given a dataset \( \mathcal{D} \), we cluster the data into \( K=100 \) clusters with centroids \( C = \{c_1, c_2, \dots, c_{100}\} \). Let \( e(x) \) denote the embedding of a code snippet \( x \). We pre-normalize all the embeddings to have magnitude 1. The distance metric we use is cosine-similarity, $d_C(x,y) = 1 - x^\top y$, where $x,y$ are both unit norm.

\paragraph{Distance to Centroid:}
For each code snippet \( x \in \mathcal{D} \), we determine the nearest centroid and compute its distance as:
\[ c_{\text{min}}(x) = \arg\min_{c_i \in C} d_C(e(x), c_i) \]
\[ d(x) = d_C(e(x), c_{\text{min}}(x)) \]

\paragraph{Cluster Size:}
For each centroid \( c_i \), we determine the size of its associated cluster, \( s(c_i) \) which equals the number of points assigned to it. For any snippet \( x \), the associated cluster size is given by \( s(c_{\text{min}}(x)) \).

\paragraph{Pruning Strategy:}
To prune the dataset, we first rank the code snippets based on their associated cluster size \( s(c_{\text{min}}(x)) \) and their distance \( d(x) \). We then prune the top \( p=20\% \) of the dataset. To interpolate between pruning based on cluster size and distance, we specify a hyperparameter $\alpha$. We then prune $\alpha p\%$ of data based on cluster size (removing smaller clusters) and $(1-\alpha) p\%$ of the remaining data based on distance (removing points further from centroids). We experiment with multiple values of $\alpha$, with full results in \cref{tab:prune_sweep}. This pruning mechanism ensures that the data points that are most similar to our synthetically corrupted data, in terms of spatial properties, are removed, thus refining \( \mathcal{D} \) to a cleaned version \( \mathcal{D}_{\text{clean}} \).

\begin{table}[]
    \centering
    \caption{Effects of different percentages of pruning from small clusters and from points far from cluster centroids.}
    \begin{tabular}{l|cccccc}
        $\alpha$ & 0.0 & 0.2 & 0.5 & 0.7 & 0.8 & 1.0 \\
        \hline
        HumanEval & 26.8\% & 22.6\% & 23.8\% & 23.8\% & 28.0\% & 23.2\% \\
        MBPP & 30.8\% & 31.6\% & 33.2\% & 31.8\% & 33.0\% & 35.0\% 
    \end{tabular}
    \label{tab:prune_sweep}
\end{table}

The pseudocode can be found at \cref{alg:pruning_weighted_ranking}.

\begin{algorithm}[h]
\caption{Embedding-guided Weighted
Pruning of Code Data}
\label{alg:pruning_weighted_ranking}
\begin{algorithmic}[1]
\Require Dataset \( \mathcal{D} \), fraction \( p \) to prune, embedding function \( e(\cdot) \), weight \( \alpha \) between [0, 1]
\Ensure Pruned dataset \( \mathcal{D}_{\text{pruned}} \)

\State Cluster \( \mathcal{D} \) into \( K \) clusters with centroids \( C = \{c_1, c_2, \dots, c_K\} \)
\State Calculate cluster sizes \( s(c_i) \) for \( c_i \in C \)

\For{each \( x \) in \( \mathcal{D} \)}
    \State \( c_{\text{min}}(x) \gets \arg\min_{c_i \in C} d_C(e(x), c_i) \) \Comment{Find closest centroid}
    \State \( d(x) \gets d_C(e(x), c_{\text{min}}(x)) \) \Comment{Compute distance to closest centroid}
\EndFor

\State Rank \( \mathcal{D} \) based on \( s(c_{\text{min}}(x)) \) in ascending order
\State \( \mathcal{D}_{\text{prune\_by\_size}} \gets \) top \( \alpha \times p \% \) of \( \mathcal{D} \) based on cluster size ranking
\State Rank remaining \( \mathcal{D} \setminus \mathcal{D}_{\text{prune\_by\_size}} \) based on \( d(x) \) in descending order
\State \( \mathcal{D}_{\text{prune\_by\_distance}} \gets \) top \( (1-\alpha) \times p \% \) of \( \mathcal{D} \setminus \mathcal{D}_{\text{prune\_by\_size}} \) based on distance ranking




\State \( \mathcal{D}_{\text{pruned}} \gets \mathcal{D} \setminus (\mathcal{D}_{\text{prune\_by\_size}} \cup \mathcal{D}_{\text{prune\_by\_distance}}) \)  \Comment{Remove the pruned data}

\Return \( \mathcal{D}_{\text{pruned}} \)

\end{algorithmic}
\end{algorithm}

\section{Evaluation}
\label{sec:app-eval}
\subsection{Metric}

The \( \text{pass@k} \) metric evaluates the functional correctness of the generated code. Specifically, a code sample is considered ``correct'' if it successfully passes a set of unit tests. For each problem, \( k \) code samples are generated, and the problem is considered ``solved'' if any of those samples pass the unit tests. The metric \( \text{pass@k} \) ultimately reports the total fraction of problems that are solved.

We define \( \text{pass@k}_i(n, c, k) \) for each \( x_{\text{test}, i} \) as:

\[
\text{pass@k}_i(n, c, k) = \begin{cases}
1.0, & \text{if } n - c < k, \\
1.0 - \prod_{j=n-c+1}^{n} \left(1 - \frac{k}{j}\right), & \text{otherwise}.
\end{cases}
\]

Here, \( n \) is the total number of samples in \( \mathcal{D}_{\text{test}} \), \( c \) is the number of correct samples for the \(i\)-th test case, and \( k \) is the value for which \( \text{pass@k}_i \) is calculated. 

\subsection{Datasets}
\label{sec:eval_data}
\paragraph{HumanEval}
HumanEval \citep{chen2021evaluating} consists of 164 hand-crafted programming tasks aimed at assessing functional correctness. The hand-crafted nature minimizes overlap with GitHub-based training data. Each task includes essential elements like function signature, docstring, and an average of 7.7 unit tests, targeting skills like language comprehension, reasoning, and algorithms.

\paragraph{MBPP: Mostly Basic Programming Problems}
MBPP \citep{austin2021program} features 974 crowd-sourced Python programs, varying from simple calculations to tasks requiring specialized knowledge. Each problem has a problem statement, a Python function, and three test cases, along with a ground-truth solution. For our experiments, we use an edited subset of 426 questions that adhere to standard Python conventions and are unambiguous. 

\begin{figure}
    \centering
    \includegraphics[width=0.5\textwidth]{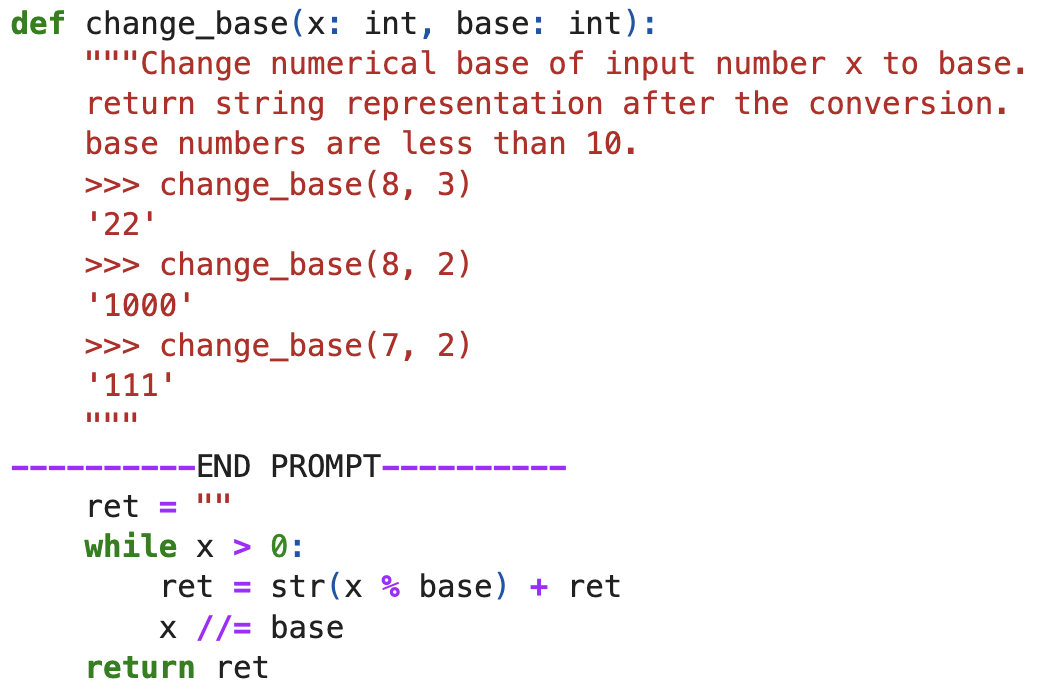}\\
    (a) HumanEval example \\
    \includegraphics[width=\textwidth]{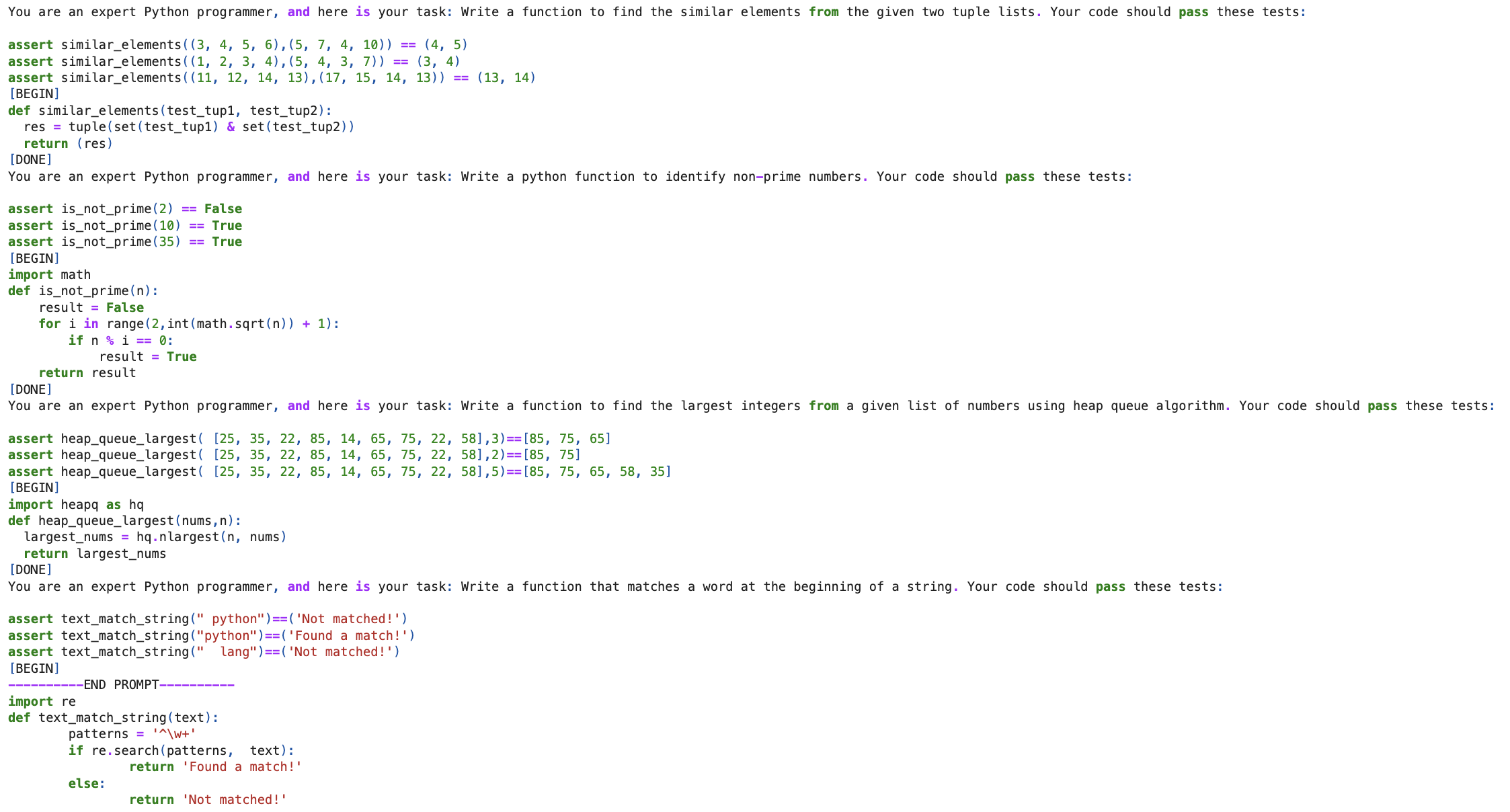} \\
    (a) MBPP example
    \caption{Example prompt and solutions for (a) HumanEval and (b) MBPP. ``END PROMPT'' is added in artificially for reader's clarity -- that line does not appear in the actual prompt or solution.}
    \label{fig:eval_example}
\end{figure}

We follow the standard procedure to evaluate models in zero-shot
on HumanEval and 3-shot on MBPP. Example prompt and answers are provided in Figure \cref{fig:eval_example}.

\subsection{Baselines}

\paragraph{SSL-prototypes} SSL prototypes \citep{sorscher2022beyond} presents a data pruning based on the underlying theory for perceptrons. This method makes three predictions relevant to broader neural network applications and benchmark dataset training. Firstly, when the initial dataset size is large, emphasizing the most challenging examples will be more helpful compared to random data pruning. Secondly, when data pruning retains a fixed fraction of the toughest examples, the outcome should exhibit power law scaling consistent with that of random pruning as the original dataset size grows. Lastly, optimizing test error over both the initial dataset size and the retained fraction can potentially produce a Pareto optimal curve that surpasses the power law scaling concerning the pruned dataset size. This is achieved by more aggressive pruning for larger initial dataset sizes. To devise a self-supervised pruning metric the SSL prototypes method employs k-means clustering within the embedding space of a pre-trained self-supervised model, such as SWaV \citep{caron2020unsupervised}. A data point's difficulty is determined by its cosine distance to the closest cluster centroid or prototype, implying that ``easy'' examples align closely with the prototype, while ``hard'' ones deviate significantly. This self-supervised prototype metric either matches or outperforms some of the best supervised metrics up to retaining 70-80\% of the data for image datasets. 

\paragraph{SemDeDup} The process of detecting perceptual duplicates might be straightforward in input space, but identifying semantic duplicates presents a unique challenge. This is primarily because semantic duplicates can be considerably different in pixel or token spaces. The SemDeDup method \citep{abbas2023semdedup}tackles this issue by employing the embedding space of a large pre-trained foundational model, which offers a semantically-rich distance metric. To detect and eliminate semantically similar entries, the algorithm first embeds each data point using foundational models, such as CLIP \citep{radford2021learning} for images and OPT \citep{zhang2022opt} for text. Subsequent clustering of these embeddings is done using k-means. Within each cluster, pairwise cosine similarities are computed, setting a threshold above which entries are flagged as semantic duplicates. Only the entry with the least cosine similarity to the cluster's centroid is retained, while the others are pruned. 

\paragraph{D4}D4 \citep{tirumala2023d4}: While working with large datasets, encountering clusters of redundant or templated text is common. Such clusters, which may not be filtered out by methods like MinHash, create dense regions in the embedding space. This density can influence clustering algorithms like k-means to allocate clusters to duplicated text, which can compromise the efficiency of methods like SSL Prototypes, where many clusters may be dominated by duplicates rather than topic-based coherence. Recognizing this, the D4 strategy was introduced. It starts by applying the SemDeDup method on the entire dataset produce a de-duplicated dataset. This pruned dataset is then clustered using K-Means. Subsequently, SSL Prototypes is applied. The resulting strategy ensures a global and local diversification of data. This method is abbreviated as D4, denoting ``Document De-Duplication and Diversification''.

\section{Inspecting the Pruned Data}

\begin{figure}
     \centering
     \begin{subfigure}[b]{\textwidth}
         \centering
         \includegraphics[width=\textwidth]{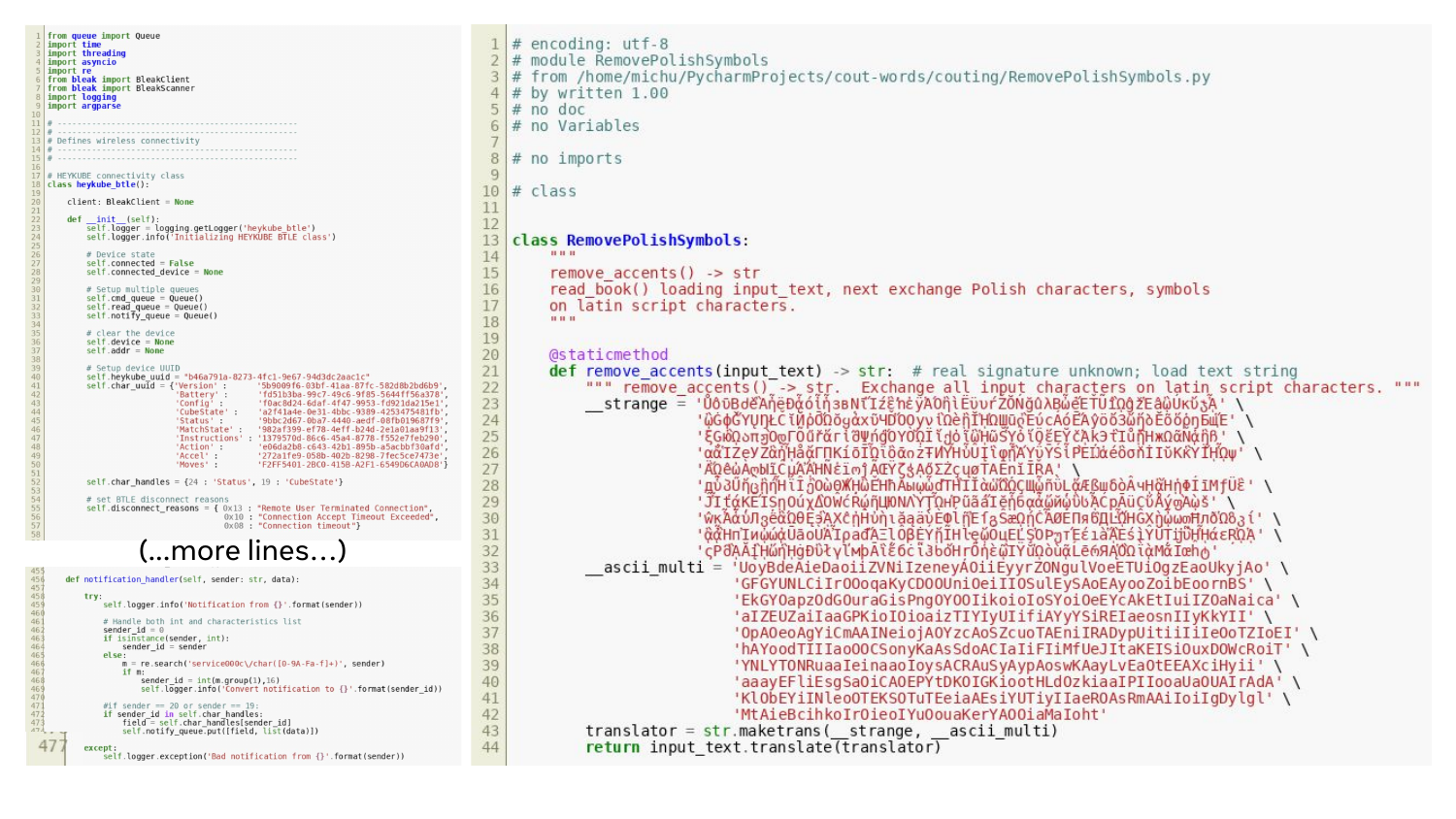}
         \caption{Top 2 examples pruned for their low similarity from the centroids. They both contain a lot of meaningless symbols.}
         \label{fig:far-to-centroids}
     \end{subfigure}
     \hfill
     \begin{subfigure}[b]{\textwidth}
         \centering
         \includegraphics[width=\textwidth]{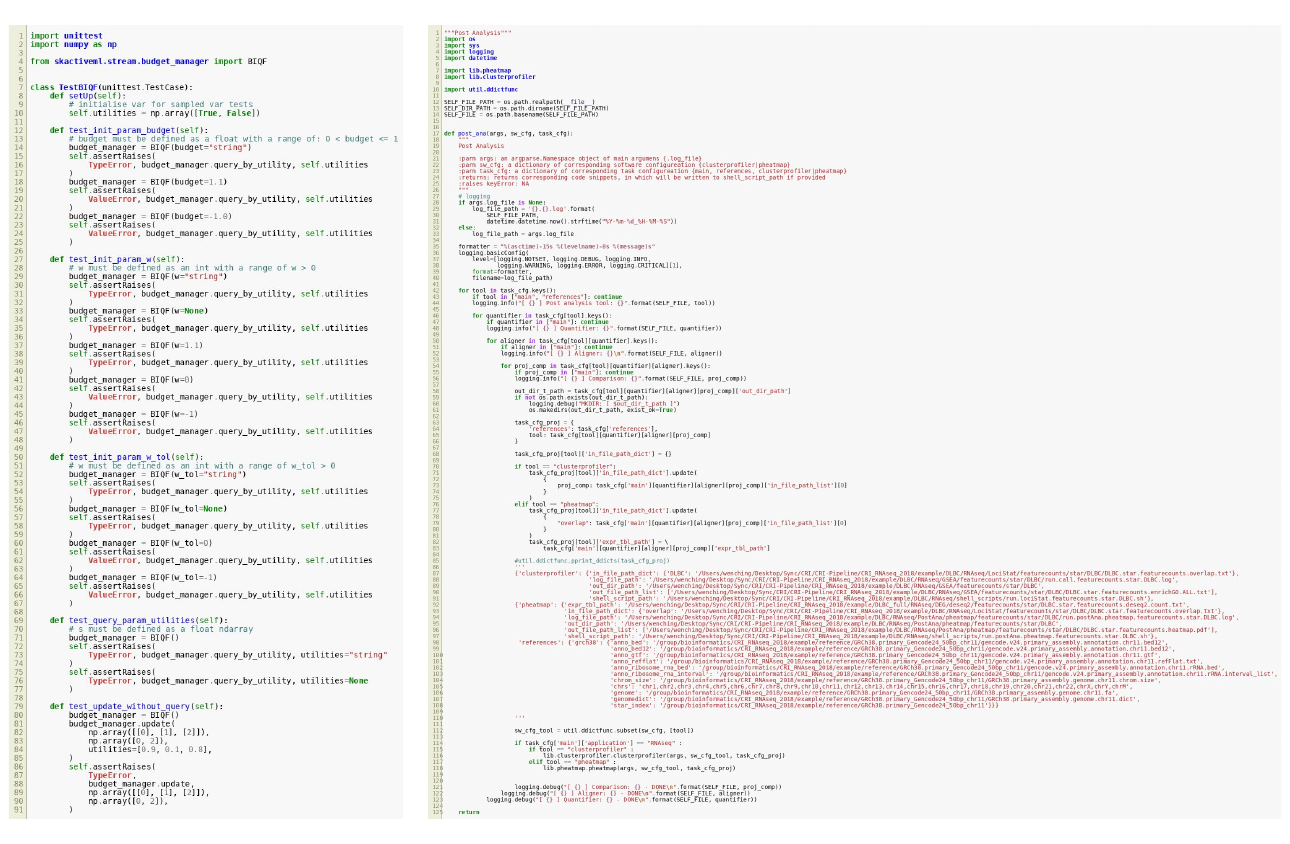}
         \caption{Top 2 examples pruned from the smallest clusters. They either repeat multiple similar functions, or contain extremely long lines with not instructive file paths.}
         \label{fig:small-clusters}
     \end{subfigure}
    \caption{Examples of pruned data points.}
    \label{fig:examples}
\end{figure}

%% file: sections/2_related.tex
\paragraph{Embedding-based Data Pruning for LLMs}
With the advent of LLMs, data quality characterization has become even more critical. SemDeDup \citep{abbas2023semdedup} exemplifies how quality embeddings can expedite LLM learning with reduced data. By removing semantic duplicates, SemDeDup enhances both training efficiency and downstream performance for LLMs. Extending this further, D4 (Document De-Duplication and Diversification) \citep{tirumala2023d4} combines SemDeDup with SSL Prototypes \citep{sorscher2022beyond} and outperforms using both SemDeDup and SSL Prototypes independently. These findings offer insights into continual model improvement beyond data scale. While these methods use embedding spaces to prune, the heuristics employed on top (pruning more ``prototypical'' examples or close by points) are hand-designed. In this work, we extend prior work by showing the promise of synthetically corrupted data as a signal for embedding space heuristics for pruning.

\paragraph{Code Generation}
Remarkable advancements have been made in recent years in the development of code-generating AI assistants \cite{roziere2023code}. These models, known as code LLMs, are crafted by training large transformer neural networks on extensive corpora of source code, empowering them to perform sophisticated code completions. They can generate code not only from surrounding code fragments but also from natural language instructions. To evaluate code LLMs, researchers use unit tests to check if the generated code behaves as expected. Two popular benchmarks for evaluating Python code generation are HumanEval \citep{chen2021evaluating} and MBPP \citep{austin2021program}, which include descriptions of functions in plain language along with corresponding sets of unit tests.
